\documentclass{article}
\usepackage{spconf,amsmath,graphicx}
\usepackage{dcolumn}% Align table columns on decimal point
\usepackage{amsmath,amsfonts}% popular packages from the American Mathematical Society
\usepackage{bm}% Bold Math package
\usepackage{multirow}
\usepackage{tabularx}
\usepackage{tikz}
\usepackage{mathtools,amssymb}
\usepackage{breqn}

\title{Unsupervised neural adaptation model based on optimal transport for spoken language identification}
\name{Xugang Lu$^{1*}$, Peng Shen$^{1}$, Yu Tsao$^{2}$, Hisashi Kawai$^1$ \thanks{*Acknowledgment: The work is supported by JSPS KAKENHI
No. 19K12035.}}
\address{$^1$National Institute of Information and Communications
Technology, Japan\\
  $^2$Research Center for Information Technology Innovation, Academic
Sinica, Taiwan}
\begin{document}
%\ninept
%
\maketitle
\begin{abstract}
Due to the mismatch of statistical distributions of acoustic speech between training and testing sets, the performance of spoken language identification (SLID) could be drastically degraded. In this paper, we propose an unsupervised neural adaptation model to deal with the distribution mismatch problem for SLID. In our model, we explicitly formulate the adaptation as to reduce the distribution discrepancy on both feature and classifier for training and testing data sets. Moreover, inspired by the strong power of the optimal transport (OT) to measure distribution discrepancy, a Wasserstein distance metric is designed in the adaptation loss. By minimizing the classification loss on the training data set with the adaptation loss on both training and testing data sets, the statistical distribution difference between training and testing domains is reduced. We carried out SLID experiments on the oriental language recognition (OLR) challenge data corpus where the training and testing data sets were collected from different conditions. Our results showed that significant improvements were achieved on the cross domain test tasks.
\end{abstract}
\section{Introduction}
\label{sec-I}
Designing a robust spoken language identification (SLID) algorithm is very important for the wide usability of multi-lingual speech applications \cite{Mabin2013,Dehak2011}. With the resurgence of deep model learning, the SLID performance has been significantly improved by current supervised deep feature and classifier learning algorithms \cite{Snyder2018,RichardsonIEEE,Ranjan2016,Moreno2016,Moreno2014,Diez2015,Fernando2017,Geng2016}. In most algorithms, there is an implicit assumption that the training and testing data sets share a similar statistical distribution property. However, due to the complex acoustic and linguistic patterns, it is often the case that testing data set and training data set are from quite different domains (e.g., different utterance durations or recording environments). An intuitive solution is to do domain adaptation, i.e., to align the statistical distribution of testing data set to match that of training data set thus to improve the performance. Although with large collected labeled testing data set, it is not difficult to obtain a domain transfer function with supervised learning algorithms, in real applications, the label information of testing data set is often unknown. Therefore, in this study, we mainly focus on a more preferable and challenge situation, i.e., unsupervised domain adaptation.

Unsupervised domain adaptation algorithms have been proposed for speaker verification, e.g., probabilistic linear discriminant analysis (PLDA) parameter adaptation  \cite{Romero2014ODS,Romero2014SLT}, feature-based correlation alignment (CORAL) \cite{LeeICASSP2019}, and feature-distribution adaptor for different domain vectors \cite{Bousquet2019}. However, in these algorithms, most of them were proposed for speaker verification under the framework of the PLDA \cite{Prince2007}. As our experiments showed that the PLDA framework does not perform well for our SLID task due to the less of discriminative power of the modeling. Instead, in most SLID algorithms, a multiple mixture of logistic regression (LR) model is used as a classifier model. Moreover, due to the complex shapes of the distributions in training and testing domains, it is difficult to guarantee the match between different domain distributions.

The purpose for domain adaptation is to reduce the domain discrepancy. Recently, optimal transport (OT) has been intensively investigated for domain adaptation in machine learning field \cite{CourtyIEEE2017}. The initial motivation for OT in machine learning is to find an optimal transport plan to convert one probability distribution shape to another shape with the least effort \cite{Peyre2018}. By finding the optimal transport, it naturally defines a distance measure between different probability distributions. Based on this property, the OT is a promising tool for domain adaptation and shape matching in image processing, classification, and segmentation \cite{CourtyIEEE2017,CourtyNIPS2017, DamodaranECCV2018}. In this paper, inspired by the OT based unsupervised adaptation \cite{CourtyIEEE2017,CourtyNIPS2017, DamodaranECCV2018}, we propose an unsupervised neural adaptation framework for cross-domain SLID tasks. Our main contributions are:
(1) We propose an unsupervised neural adaptation model for SLID to deal with domain mismatch problem. In the model, we explicitly formulate the adaptation in transformed feature space and classifier space in order to reduce the probability distribution discrepancy between source and target domains.
(2) We coincide the OT distance metric in measuring the probability distribution discrepancy, and integrate it into the network optimization in order to learn the adaptation model parameters. Based on the adaptation model, significant improvements were obtained.
%The remainder of the paper is organized as follows. Section \ref{sec_frm} introduces the background and fundamental theory of . Section \ref{sec-III} describes the implementation details of . Section \ref{sec-IV} presents the SLID experiments and results based on the proposed framework by analyzing the contribution of the CSA model in detail. Section \ref{sec-V} presents the discussion of the results and conclusion of the study.
\section{Proposed adaptation framework}
\label{sect_frm}
\begin{figure}[tb]
\centering
\includegraphics[width=6cm, height=5.5cm]{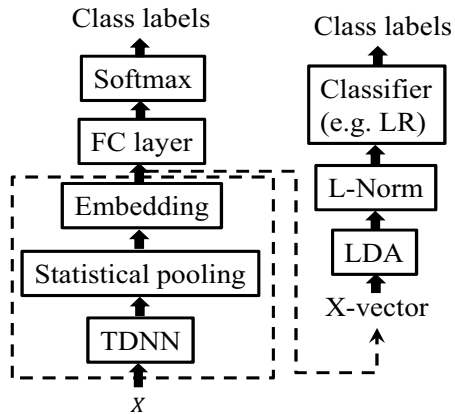}
\caption{Conventional framework for SLID based on X-vector extraction and classifier models}
\label{figConv}
\end{figure}

%\subsection{State of the art model architecture for SLID}
Conventional state of the art model for SLID is built based on X-vector representations as illustrated in Fig. \ref{figConv}.
In this figure, the left part is the model for X-vector extraction which is based on time delay neural network (TDNN) \cite{Snyder2018}, FC layer denotes full connected layer. The back-end model on the  X-vector is showed in the right part of Fig. \ref{figConv}. ``LDA" means linear discriminant analysis (LDA), ``L-Norm" is length normalization for feature normalization. Based on this normalized feature, a discriminative classifier, e.g., logistic regression (LR) is applied in the final classification. Although this model architecture in Fig. \ref{figConv} could relieve the mismatch problem in domain changes in some degree, it is still not enough when the difference between training and testing distributions is large. In this study, we explicitly apply domain adaptation process based on this model framework.
\subsection{Cross domain adaptation}
\label{sect-cross}
Given a source domain data set $D_s  = \left\{ {\left( {{\bf x}_i^s ,{\bf y}_i^s } \right)} \right\}_{i = 1,..,N}$, and target domain data set $D_t  = \left\{ {\left( {{\bf x}_i^t ,{\bf y}_i^t } \right)} \right\}_{i = 1,..,M}$ (in real situations, the target label information is unknown, and predicted label will be used in unsupervised adaptation as showed in section \ref{subsect_NNOT}). Due to domain changes (e.g. recording channels), there exists domain discrepancy, i.e., $p_s \left( {{\bf x},{\bf y}} \right) \ne p_t \left( {{\bf x},{\bf y}} \right)$. The purpose for domain adaptation is to reduce this discrepancy.
\subsubsection{Problem formulation}
In order to remove the domain distribution difference, we want to find a transform $\phi \left(  \cdot  \right)$ by which $p_s \left( {\phi \left( {\bf x} \right),{\bf y}} \right) \approx p_t \left( {\phi \left( {\bf x} \right),{\bf y}} \right)$. Based on the Bayesian theory, the joint distribution of feature and label in the transformed space is formulated as:
{\small
\begin{equation}
p\left( {\phi \left( {\bf x} \right),{\bf y}} \right) = p\left( {\phi \left( {\bf x} \right)} \right)p\left( {{\bf y}|\phi \left( {\bf x} \right)} \right).
\end{equation}
}
Correspondingly, the joint distributions in source and target domains should be approximated. The approximation can be explicitly decomposed to the latent feature approximation (marginal distribution) and classifier approximation (conditional distribution) as:
{\small
\begin{equation}
\begin{array}{l}
 p_s \left( {\phi \left( {{\bf x}^s } \right)} \right) \approx p_t \left( {\phi \left( {{\bf x}^t } \right)} \right) \\
 p_s \left( {{\bf y}^s |\phi \left( {{\bf x}^s } \right)} \right) \approx p_t \left( {{\bf y}^t |\phi \left( {{\bf x}^t } \right)} \right) \\
 \end{array}
\label{eq_joint}
\end{equation}
}
From Eq. (\ref{eq_joint}), we can see that $p_s ( {\phi ( {{\bf x}^s } )} )$ and $p_t ( {\phi ( {{\bf x}^t } )} )$ are the latent feature distributions, and $p_s \left( {{\bf y}^s |\phi \left( {{\bf x}^s } \right)} \right)$ and $p_t \left( {{\bf y}^t |\phi \left( {{\bf x}^t } \right)} \right)$ are the posterior distributions given the latent features for source and target samples. The intuitive way is to minimize the two distribution differences. In order to minimize the distribution difference, a distance measure between the different distributions should be defined.
\subsubsection{Optimal transport distance}
 Optimal transport (OT) distance (also known as  Wasserstein distance) is an efficient measure of distance between probability distributions with consideration of the geometric shape of the distributions \cite{CourtyIEEE2017,CourtyNIPS2017}. For two distributions $p_1$ and $p_2$, the OT distance is defined as:
{\small
\begin{equation}
L_{{\rm OT}} ( {p_1 ,p_2 } ) = \mathop {\inf }\limits_{\gamma  \in \prod {( {p_1 ,p_2 } )} } \int\limits_{\Re ^1  \times \Re ^2 } {C( {{\bf z}_1 ,{\bf z}_2 } )d\gamma ( {{\bf z}_1 ,{\bf z}_2 } )},
\label{eq_ot}
\end{equation}
}
where `$\inf$'  means infimum, i.e., the greatest lower bound of the integration set, ${\gamma  \in \prod {\left( {p_1 ,p_2 } \right)} }$ is the transport plan (or coupling) between the two probability distributions, and ${C\left( {{\bf z}_1 ,{\bf z}_2 } \right)}$ is the cost between sample points ${\bf z}_1$ and ${\bf z}_2$, where $\left( {{\bf z}_1 ,{\bf z}_2 } \right)$ is a sample pair from a compact space $\Re ^1  \times \Re ^2$ with marginal distributions $ p_1$ and $p_2$, respectively. From Eq. (\ref{eq_ot}), we can see that the OT distance can be regarded as a weighted pair-wised distance between samples from two distributions. The initial OT based adaptation is applied for finding a marginal latent feature space, and later it is modified for joint adaptation framework, i.e., both feature (marginal distribution) and classifier (conditional distribution) are adapted in image classification \cite{CourtyIEEE2017, CourtyNIPS2017, DamodaranECCV2018}. In this study, we coincide the OT loss in a neural network optimization for SLID.
 \subsection{Neural network for domain adaptation with OT}
 \label{subsect_NNOT}
\begin{figure}[tb]
\centering
\includegraphics[width=8cm, height=6cm]{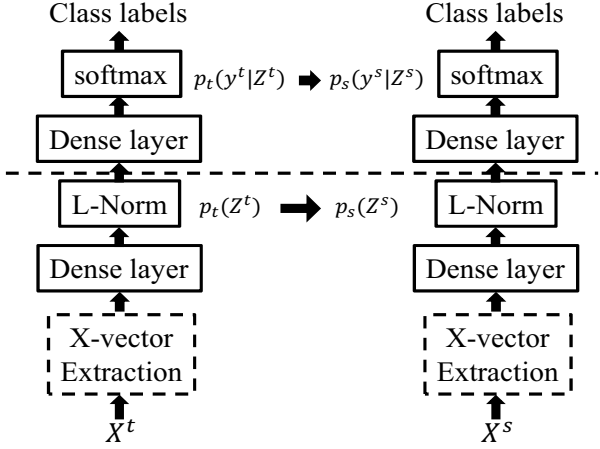}
\caption{Joint distribution adaptation neural network for SLID}
\label{fig_JBFrm}
\end{figure}

In a neural network based classification model, it can be explicitly regarded as a composition of two modules, i.e., feature extraction module and classifier module defined as:
{\small
\begin{equation}
y\left( {\bf x} \right) = f\left( {{\bf x};\theta _g ,\theta _h } \right) = g \circ h\left( {\bf x} \right),
\label{eq_comp}
\end{equation}
}
where $h\left(  \cdot  \right)$ and $g\left(  \cdot  \right)$ are the feature extraction and classifier transforms with parameter sets ${\theta _g }$ and ${\theta _h }$, respectively. The adaptation process could be explicitly applied to the outputs of the two modules. Connecting the domain adaptation in feature and classifier, the proposed adaptation network architecture is illustrated in Fig. \ref{fig_JBFrm}.
In this figure, we suppose that the latent feature space, ${\bf z} = \phi \left( {\bf x} \right) = h\left( {\bf x} \right)$ is via a series of transforms: X-vector extraction directly adopted from model architecture in Fig. \ref{figConv}, a linear dense layer transform which simulates the LDA function as used in Fig. \ref{figConv}, a length normalization (L-Norm) before the feature is input to the classifier layer.  The classifier module is configured with a linear dense layer followed by a softmax activation function which outputs class-wise probability values.  With reference to the explanation in section \ref{sect-cross}, the adaptation is defined to minimize the following loss as:
{\small
\begin{equation}
L_{{\rm adapt}}\! ( {{\bf x}^s  \! , {\bf y}^s \! ; {\bf x}^t ,{\bf y}^t } ) \! = \!  \alpha L_{{\rm fea}} ( {\phi (\! {{\bf x}^s } ),\phi ( \!{{\bf x}^t } )} ) \! + \!  \beta L_{{\rm cls}} ( {{\bf y}^s ,{\bf y}^t } ),
\label{eq_Ladapt}
\end{equation}
}
where $L_{{\rm fea}} ( {\phi ( {{\bf x}^s } ),\phi ( {{\bf x}^t } )} )$ is used to measure the feature distribution difference (loss) in a latent space, $L_{{\rm cls}} \left( {{\bf y}^s ,{\bf y}^t } \right)$ for classifier distribution difference (loss), $\alpha$ and $\beta$ are weighting coefficients for feature and classifier adaptation losses (simple Euclidian distance is used in this paper). In real situations, label in target domain is usually unknown. Instead, an estimation ${\bf \hat y}^t  = f\left( {{\bf x}^t } \right) = g \circ h\left( {{\bf x}^t } \right)$ is used for target label. As an unsupervised adaptation, the alignment between source and target domains is unknown, therefore, in the definition with different samples in source and targets domains, the loss is a pair-wised loss which is defined as OT distance:
{\small
\begin{equation}
L_{{\rm OT}} ( {p_s ,p_t } ) = \mathop {\min }\limits_{\gamma  \in \prod {( {p_s ,p_t } )} } \sum\limits_{i,j} {L_{{\rm adapt}}( {{\bf z}_i^s ;{\bf z}_j^t } )} \gamma ( {{\bf z}_i^s ;{\bf z}_j^t } ),
\label{eq_OTAdpt}
\end{equation}
}
where ${\bf z}_i^s  \in \left\{ {{\bf x}_i^s ,{\bf y}_i^s } \right\}$ and ${\bf z}_j^t  \in \left\{ {{\bf x}_j^t ,{\bf y}_j^t } \right\}$ are samples from source and target domains ($i$ and $j$ are sample indexes), respectively. Eq. (\ref{eq_OTAdpt}) is used to find the optimal transport plan matrix $\gamma$ by which the adaptation loss could be estimated.

Besides the adaptation loss, the classification loss in source domain is defined as the multi-class cross entropy as:
{\small
\begin{equation}
L_{{\rm CE}}^s \left( {{\bf y}_i^s ,{\bf \hat y}_i^s } \right) =  - \sum\limits_{j = 1}^c {y_{i,j}^s \log \hat y_{i,j}^s },
\end{equation}
}
where ${{\bf \hat y}_i^s }$ is the estimated label in source domain as ${\bf \hat y}_i^s  = f\left( {{\bf x}_i^s } \right) = g \circ h\left( {{\bf x}_i^s } \right)$ ($i$ as sample index, and $c$ is the number of class).
Therefore, the total loss including the adaptation loss and source domain classification loss as:
{\small
\begin{equation}
L_{\rm T}  = \mathop {\min }\limits_{\gamma ,\theta _g ,\theta _h } \left( {\sum\limits_i {L_{{\rm CE}}^s \left( {{\bf y}_i^s ,{\bf \hat y}_i^s } \right)}  + \lambda L_{{\rm OT}} \left( {p_s ,p_t } \right)} \right).
\label{eq_TLoss}
\end{equation}
}
In the above equation, there are three sets of parameters to be optimized, i.e.,  optimal transport plan matrix $\gamma$ as defined in Eq. (\ref{eq_OTAdpt}), feature transform related parameters ${\theta _g }$, and classifier transform related parameters ${\theta _h }$ as defined in Eq. (\ref{eq_comp}). As defined in Eqs. (\ref{eq_comp}), (\ref{eq_Ladapt}) and (\ref{eq_OTAdpt}), the estimation of optimal transport plan matrix needs feature extraction and class prediction for both source and target domains, while the optimal feature extraction and prediction depend on the optimal transport plan. The optimization could be put into an expectation-maximization (EM) like framework with mini-batch sampling of source and target samples as introduced in \cite{CourtyNIPS2017,DamodaranECCV2018}
\section{Experiments}
\label{sect_exp}
\subsection{Experimental conditions}
Experiments on SLID task are carried out based on data corpus from Oriental Language Recognition (OLR) 2020 Challenge \cite{WD2019,WD2020,OLR2020}. In training, there are around 110 k utterances ( more than 100 hours),  from 10 languages. For test sets, two tasks are performed, i.e., short utterance LID test, and cross channel LID test (task 1 and task 2 as used in OLR 2019 challenge).  Task 1 test set includes utterances from the same 10 languages of the training set (1.8 k utterances for each), but the utterance duration is short (1 s). Task 2 includes only 6 languages, also with 1.8 k utterances for each. But the utterances were recorded in wild environments which are quite different from those of for training data set. In task 2, there is one development set and one test set. In this paper, both are used as independent test sets for task 2. In order to measure the quality of the classification and adaptation models, two evaluation metrics are adopted by considering the missing and false alarm probabilities for target and nontarget language pairs, i.e., equal error rate (EER), and average performance cost (Cavg) as defined in \cite{WD2020}.
%\vspace{-2mm}
%\subsection{Language embedding based on x-vector}

Similarly as used in speaker embedding \cite{Snyder2018}, the language embedding model is with an extended TDNN architecture as implemented in the baseline model \cite{OLR2020}. For improving the robustness, data augmentation techniques are applied \cite{OLR2020}. Input features for training the language embedding model are 30 dimensional MFCCs with 36 Mel band bins. And the MFCCs are extracted with 25 ms frame length and 10 ms frame shift. The final extracted x-vectors are with 512 dimensions. Two baseline systems are constructed, one is conventional baseline system (illustrated in Fig. \ref{figConv}) as implemented in \cite{OLR2020}. The other is our proposed neural back-end model as illustrated in Fig. \ref{fig_JBFrm}. From this figure, we can see that the dimension reduction (200 dimensions) and length normalization are integrated in the optimization process. As our experiments showed that the performances of these two baseline systems are comparable, we will not show the results of the conventional baseline system due to limited of space. The adaptation is based on our proposed neural back-end system, and the adaptation process is optimized using the Adam algorithm with an initial learning rate of $0.001$ \cite{Adam}, mini-batch size was 128.
%\begin{figure*}[tb]
%\centering
%\includegraphics[width=10cm, height=3.5cm]{./Figures_LID/Fig_TSNE_train_test12_clusterMM.eps}
%\vspace{-2mm}
%\caption{Language cluster distributions based on the TSNE for: training set (a), test set for task 1 (b), and test set for task 2 (c).}
%\vspace{-3mm}
%\label{figBJB}
%\end{figure*}
\begin{figure}[tb]
\centering
\includegraphics[width=8cm, height=4cm]{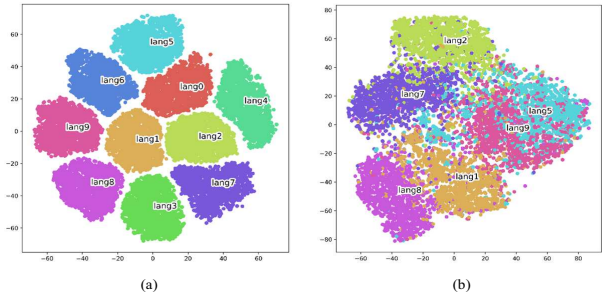}
\caption{Language cluster distributions based on the TSNE for: training data set (a), test data set for task 2 (b).}
\label{figBJB}
\end{figure}
%\vspace{-4mm}
\subsection{Results}
%\subsubsection{Baseline performance}
We first visually check the cluster distributions of the language embedding based on X-vector for both training and test data sets. The language cluster distributions are showed based on the t-Distributed Stochastic Neighbor Embedding (TSNE) \cite{Maaten2008} in Fig. \ref{figBJB}. In this figure, clusters with different colors are distributions of samples (utterances) from languages corresponding to: `Kazakh': lang0, `Tibetan': lang1, `Uyghur': lang2, `Cantonese': lang3, `Indonesian': lang4, `Japanese': lang5, `Korean': lang6, `Russian': lang7, `Vietnamese': lang8, `Mandarin': lang9. From this figure, we can see that with X-vector representation, the speaker clusters are clearly discriminated for training data set (\ref{figBJB}-a). However, for test set in task 2, there are large overlaps between different language clusters (\ref{figBJB}-b). Based on the X-vector, the baseline performance results are showed in table \ref{tab12}.
\begin{table}[tb]
\centering
\caption{Baseline performance.}
\begin{tabular}{|c||c||c|}
\hline
 Test sets&EER\% &Cavg\\
\hline
\hline
Task1\_test  &10.17&0.0978\\
\hline
Task2\_dev  &27.47&0.2746\\
\hline
Task2\_test  &27.35&0.2858\\
\hline
\end{tabular}
\label{tab12}
\end{table}
\begin{table}[tb]
\centering
\caption{Adaptation performance.}
\begin{tabular}{|c||c||c|}
\hline
 Test sets&EER\% &Cavg\\
\hline
\hline
Task1\_test  &7.11&0.0701\\
\hline
Task2\_dev  &18.60&0.1761\\
\hline
Task2\_test  &15.88&0.1583\\
\hline
\end{tabular}
\label{tab2}
\vspace{4mm}
\end{table}

In this table, Task1\_test is for short utterance LID, Task2\_dev and Task2\_test are for cross-channel LID. From these results, we can see that, the mismatch between training and test sets in task 2 is much more serious than that in task 1.

\begin{figure}[tb]
\centering
\includegraphics[width=8cm, height=4cm]{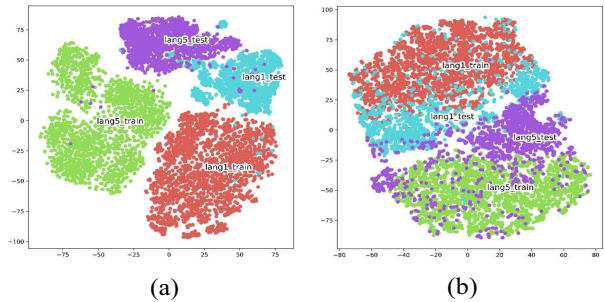}
%\vspace{-4mm}
\caption{Language cluster distributions based on the TSNE for Task2\_test: before adaptation (a), and after adaptation (b).}
\label{figJB}
\end{figure}

Based on our proposed unsupervised adaptation learning, we expect that the mismatch between training and testing is reduced. We also visually check the effect of unsupervised adaptation on language cluster distributions in Fig. \ref{figJB}.

In this figure, only two languages are showed. In Fig. \ref{figJB}-a, due to the difference of recording channels, the clusters belonging to the same language is separated (pairs of lang5\_train vs. lang5\_test, and lang1\_train vs. lang1\_test). Since the classifier is designed  based on the training data set, it is not strange that the performance of the baseline system on testing data set is degraded. After adaptation, as showed in \ref{figJB}-b, the clusters of testing data set are pushed to be overlapped with those of training data set for the same language. The recognition performance results are showed in table \ref{tab2} (with $\alpha  = 0.1,\beta  = 0.0003,\lambda  = 1.0$ as defined in Eqs. (\ref{eq_Ladapt}), (\ref{eq_TLoss}) ). Comparing the results in tables \ref{tab12} and \ref{tab2}, we can see that with our proposed unsupervised adaptation learning, the performance for both task 1 and task 2 are improved with a large margin.

As seen from Eqs. (\ref{eq_Ladapt}) and (\ref{eq_TLoss}) , with different weighting coefficients of the hyper-parameters, the adaptations for feature and classifier are different. We test the performance on test set of task 2 with variations of $\alpha$ and $\beta$ ($\lambda  = 1.0$ is fixed), the results are showed in table \ref{tab3}. The best result on this task in OLR2019 challenge (fusion from multiple systems) is also list for comparison \cite{OLR2019}. From this table, we can see that the adaptation on feature distribution is more important and effective than adaptation on the label distribution. Even without the label distribution adaptation, the performance is fairly good. However, without feature adaptation, only adapting the classifier almost does not take any effect.
\begin{table}[tb]
\centering
\caption{Effect of hyper-parameters on adaptation performance for test set of task 2.}
\begin{tabular}{|c||c||c|}
\hline
 Adaptation hyper-parameters&EER\% &Cavg\\
\hline
$\alpha  = 0.1, \beta  = 0.0$  &16.26&0.1678\\
\hline
$\alpha  = 0.1,\beta  = 0.00001$  &16.09&0.1661\\
\hline
$\alpha  = 0.1,\beta  = 0.0001$  &\textbf{15.98}&\textbf{0.1620}\\
\hline
$\alpha  = 0.1,\beta  = 0.001$  &16.46&0.1646\\
\hline
$\alpha  = 0.1,\beta  = 0.01$  &16.66&0.1726\\
\hline
$\alpha  = 0.1,\beta  = 0.1$  &21.99&0.2181\\
\hline
\hline
$\alpha  = 0.001, \beta  = 0.0001$  &25.69&0.2741\\
\hline
$\alpha  = 0.01,\beta  = 0.0001$  &22.54&0.2298\\
\hline
$\alpha  = 0.1,\beta  = 0.0001$  &\textbf{15.98}&\textbf{0.1620}\\
\hline
$\alpha  = 1.0,\beta  = 0.0001$  &17.50&0.1793\\
\hline
$\alpha  = 10.0,\beta  = 0.0001$  &25.56&0.2513\\
\hline
\hline
\emph{Best result in OLR2019} \cite{OLR2019}  &\emph{20.24}&\emph{0.2008}\\
\hline
\end{tabular}
%\vspace{-3mm}
\label{tab3}
\end{table}
\section{Discussion and conclusion}
\label{conclude}
In this paper, following the basic domain adaptation theory, a joint distribution adaptation neural network model was proposed for SLID. In our model, besides the classification loss for source domain data, we adopted an OT distance metric to measure the adaptation loss between source and target data samples. Experimental results showed that the adaptation effectively improved the performance. In this paper, we have not investigated the adaptation framework with different network architectures. In addition, concerning the adaptation loss, other probability discrepancy measurement metrics could also be integrated in current framework during model optimization. All these studies remain as our future work.
\bibliographystyle{IEEEbib}
%\bibliography{strings,refs}

\end{document}